\documentclass[letterpaper, 10 pt, conference]{ieeeconf}  

\IEEEoverridecommandlockouts                              

\usepackage{amsmath,amsfonts}
\usepackage{algorithm}
\usepackage{array}
\usepackage{textcomp}
\usepackage{stfloats}
\usepackage{url}
\usepackage{verbatim}
\usepackage{graphicx}
\usepackage{cite}
\usepackage{booktabs}
\usepackage{multirow}
\usepackage{color}
\usepackage[table]{xcolor}
\usepackage{nicematrix}
\usepackage{threeparttable}
\usepackage{makecell}
\usepackage{bbding}
\usepackage{ulem}
\usepackage{hhline}
\newcolumntype{L}[1]{>{\raggedright\let\newline\\\arraybackslash\hspace{0pt}}m{#1}}
\newcolumntype{C}[1]{>{\centering\let\newline\\\arraybackslash\hspace{0pt}}m{#1}}
\newcolumntype{R}[1]{>{\raggedleft\let\newline\\\arraybackslash\hspace{0pt}}m{#1}}

\definecolor{myBlue}{HTML}{3ba9db}
\definecolor{myRed}{HTML}{E7DBCC}
\definecolor{myYellow}{HTML}{F1F4AE}

\usepackage{calligra}
\usepackage{algorithmicx}
\usepackage{algpseudocode} 
\usepackage{amsmath} 

\errorcontextlines\maxdimen

\makeatletter
\newcommand*{\algrule}[1][\algorithmicindent]{\makebox[#1][l]{\hspace*{.5em}\thealgruleextra\vrule height \thealgruleheight depth \thealgruledepth}}%
\newcommand*{\thealgruleextra}{}
\newcommand*{\thealgruleheight}{.75\baselineskip}
\newcommand*{\thealgruledepth}{.25\baselineskip}

\newcount\ALG@printindent@tempcnta
\def\ALG@printindent{%
	\ifnum \theALG@nested>0
	\ifx\ALG@text\ALG@x@notext
	\else
	\unskip
	\addvspace{-1pt}
	\ALG@printindent@tempcnta=1
	\loop
	\algrule[\csname ALG@ind@\the\ALG@printindent@tempcnta\endcsname]%
	\advance \ALG@printindent@tempcnta 1
	\ifnum \ALG@printindent@tempcnta<\numexpr\theALG@nested+1\relax
	\repeat
	\fi
	\fi
}%
\usepackage{etoolbox}
\patchcmd{\ALG@doentity}{\noindent\hskip\ALG@tlm}{\ALG@printindent}{}{\errmessage{failed to patch}}
\makeatother

\newbox\statebox
\newcommand{\myState}[1]{%
	\setbox\statebox=\vbox{#1}%
	\edef\thealgruleheight{\dimexpr \the\ht\statebox+1pt\relax}%
	\edef\thealgruledepth{\dimexpr \the\dp\statebox+1pt\relax}%
	\ifdim\thealgruleheight<.75\baselineskip
	\def\thealgruleheight{\dimexpr .75\baselineskip+1pt\relax}%
	\fi
	\ifdim\thealgruledepth<.25\baselineskip
	\def\thealgruledepth{\dimexpr .25\baselineskip+1pt\relax}%
	\fi
	\State #1%
	\def\thealgruleheight{\dimexpr .75\baselineskip+1pt\relax}%
	\def\thealgruledepth{\dimexpr .25\baselineskip+1pt\relax}%
}

\title{\LARGE \bf
Interactive Spatiotemporal Token Attention Network for \protect\\ Skeleton-based General Interactive Action Recognition
}

\author{Yuhang Wen$^{1}$, Zixuan Tang$^{1}$, Yunsheng Pang$^{2}$, Beichen Ding$^{1*}$, Mengyuan Liu$^{3*}$
\thanks{$^{*}$Corresponding authors: {\tt\small dingbch@mail.sysu.edu.cn} (Beichen Ding) and {\tt\small nkliuyifang@gmail.com} (Mengyuan Liu)}%
\thanks{$^{1}$Yuhang Wen, Zixuan Tang and Beichen Ding are with Sun Yat-sen University, Shenzhen 518107, China.}%
\thanks{$^{2}$Yunsheng Pang is with Tencent Technology (Shenzhen) Co., Ltd., China.}%
\thanks{$^{3}$Mengyuan Liu is with the Key Laboratory of Machine Perception, Shenzhen Graduate School, Peking University, Shenzhen, China.}%
}

\begin{document}

\maketitle
\thispagestyle{empty}
\pagestyle{empty}

\begin{abstract}
Recognizing interactive action plays an important role in human-robot interaction and collaboration. Previous methods use late fusion and co-attention mechanism to capture interactive relations, which have limited learning capability or inefficiency to adapt to more interacting entities. With assumption that priors of each entity are already known, they also lack evaluations on a more general setting addressing the diversity of subjects. To address these problems, we propose an Interactive Spatiotemporal Token Attention Network (ISTA-Net), which simultaneously model spatial, temporal, and interactive relations. Specifically, our network contains a tokenizer to partition Interactive Spatiotemporal Tokens (ISTs), which is a unified way to represent motions of multiple diverse entities. By extending the entity dimension, ISTs provide better interactive representations. To jointly learn along three dimensions in ISTs, multi-head self-attention blocks integrated with 3D convolutions are designed to capture inter-token correlations. When modeling correlations, a strict entity ordering is usually irrelevant for recognizing interactive actions. To this end, Entity Rearrangement is proposed to eliminate the orderliness in ISTs for interchangeable entities. Extensive experiments on four datasets verify the effectiveness of ISTA-Net by outperforming state-of-the-art methods. Our code is publicly available at \url{https://github.com/Necolizer/ISTA-Net}.
\end{abstract}

\section{INTRODUCTION}

Interactive action recognition is a crucial yet challenging task in computer vision and physical human-robot interaction\cite{9636389,9636110,zheng2017image,DBLP:conf/iros/XingB22}, with a wide range of applications like assistive household robots\cite{9636381} and interactive mechanical arms\cite{7554295}. These smart assistants should understand interactive motion patterns and the intents behind actions to ensure safe and reliable human-robot collaboration\cite{9636107,9636553}.

\begin{figure}[t]
    \begin{center}
    \includegraphics[width=8cm]{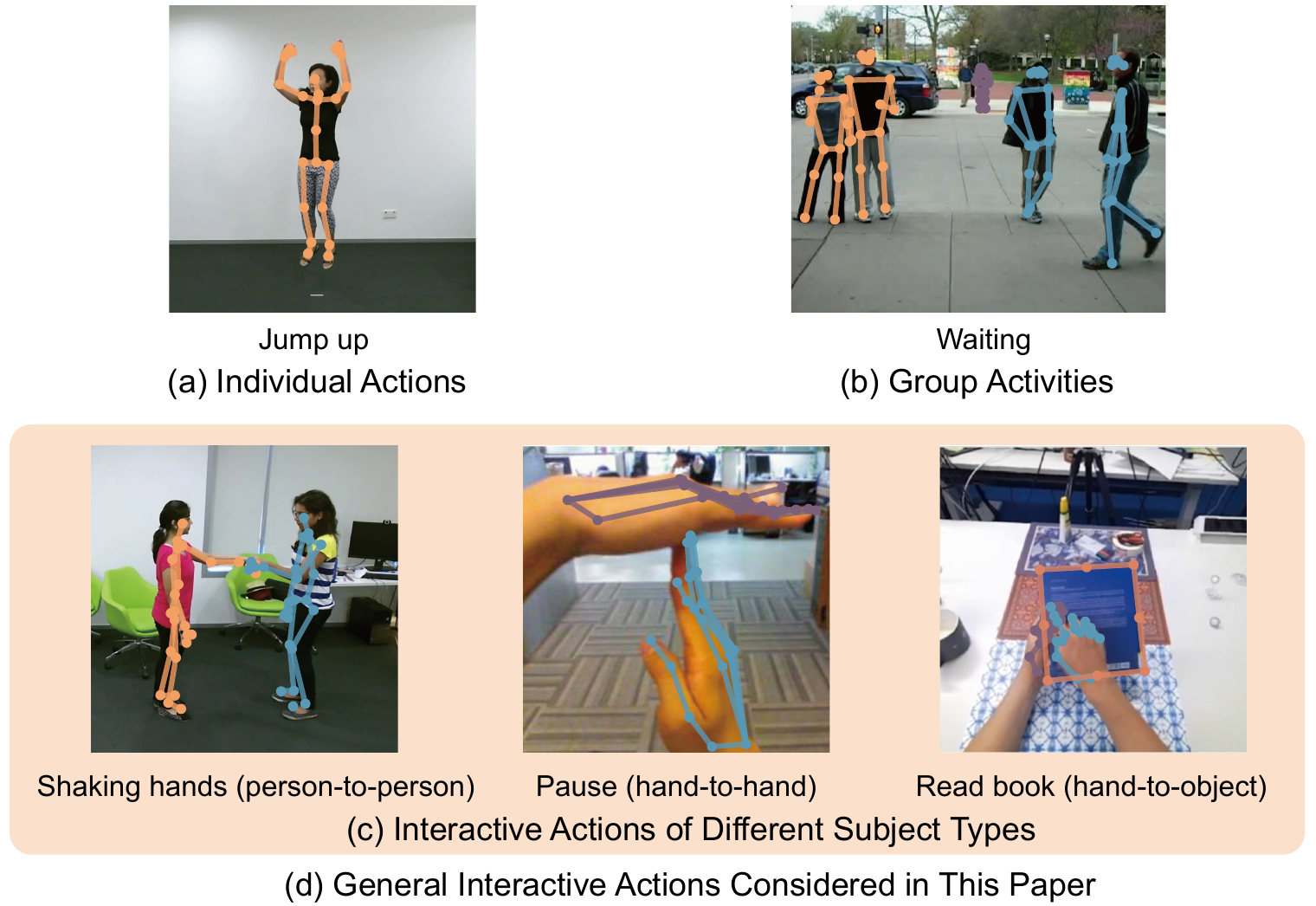}
    \end{center}
    \vspace{-1.2em}
    \caption{Examples of individual actions (a)\cite{NTU120}, group activities (b)\cite{cad2009} and interactive actions (c)\cite{NTU120,8299578,H2O_TA-GCN2021}. (a) Sequences of single pose could fully depict the action \textit{Jump Up}. (b) Group activity \textit{Waiting} is annotated regardless of the pedestrians. (c) Each entity is an integral part of the interactive action. Previous methods focus on one type of these interactions. (d) In this paper, we evaluate on general interactive action recognition task, which addresses the diversity of interacting subjects.}
    \label{InteractiveActons}
    \vspace{-1.7em}
\end{figure}

An interactive action is a purposeful behavior that involves the interdependent physical dynamics of multiple entities. The indivisibility of interdependent entities distinguish interactive actions from \textit{individual actions} and \textit{group activities}. Individual actions (Fig. \ref{InteractiveActons} (a)) are concerned with motions of a single subject. Group activities (Fig. \ref{InteractiveActons} (b)) are events concluded or abstracted from common goals of actions, which contain considerable irrelevant and noisy individual actions. In contrast, each subject in an interactive action is indispensable to illustrate the full semantic meaning. Types of interactive actions includes person-to-person, hand-to-hand, hand-to-object. Diverse interacting entities have distinct physical structures and interaction patterns, leading to the complexity and variability when modeling interactions. 

Studies on modeling interactive actions have emerged in recent years\cite{H+O2019, H2O_TA-GCN2021, LSTM-IRN2022, igformer2022}, but they study only one specific type of interactions depicted in Fig. \ref{InteractiveActons} (c). They also assume that prior knowledge of the physical connections within each interacting subject is already known and remains fixed. Therefore, these methods lack evaluations in a general setting addressing the diversity of interacting entities. \uline{In this paper, we focus on a general interactive action recognition task, which is a generalization from the subject-type-specific ones, as shown in Fig. \ref{InteractiveActons} (d).} Moreover, previous designs have limitations when capturing interactive relations. Late fusion offers a simplistic approach for modeling interactive relations, but it lacks the capacity to handle complex interactions. On the other hand, expanding the co-attention architecture to accommodate more than two interacting entities is inefficient, due to the increase in the number of calculations required for pair-wise co-attention scores, as the number of entities increases. Therefore, an important question arises: \uline{How to jointly learn the spatial, temporal, and interactive relations of diverse interacting subjects?}

To answer this question, we simultaneously model entity, temporal and spatial relations between interacting entities with an Interactive Spatiotemporal Token Attention Network (ISTA-Net), whose core component is the Interactive Spatiotemporal Tokenization. 3D Interactive Spatiotemporal Tokens (ISTs) can be generated by this tokenization, which is a unified way to represent motions of multiple diverse entities. To learn inter-token correlations, we integrate 3D convolutions with self-attention and design Token Self-Attention (TSA) Blocks. Moreover, the ordering of unordered entities in ISTs is unnecessary for modeling correlations. We propose Entity Rearrangement (ER) to solve this problem.

The main contributions of this paper are as follows:
\begin{itemize}
\item{We propose an Interactive Spatiotemporal Token Attention Network to solve the general interactive action recognition task, which does not require prior knowledge on subject's physical structure.}
\item{Specifically, we present Interactive Spatiotemporal Tokens to fuse three dimensional interactive spatiotemporal features, for effectively representing spatiotemporal interactions for diverse entities. We present Token Self-Attention Blocks for better capturing correlations of different interactive features. Moreover, Entity rearrangement is proposed to ensure inherent permutation invariance for unordered entities in ISTs.}
\item{Extensive experiments on NTU RGB+D 120, SBU-Kinect-Interaction, H2O and Assembly101 datasets consistently verify the effectiveness of our method, which outperforms most interactive action recognition methods. Our code is publicly available.}
\end{itemize}

\begin{figure*}[t]
    \begin{center}
    \includegraphics[width=17cm]{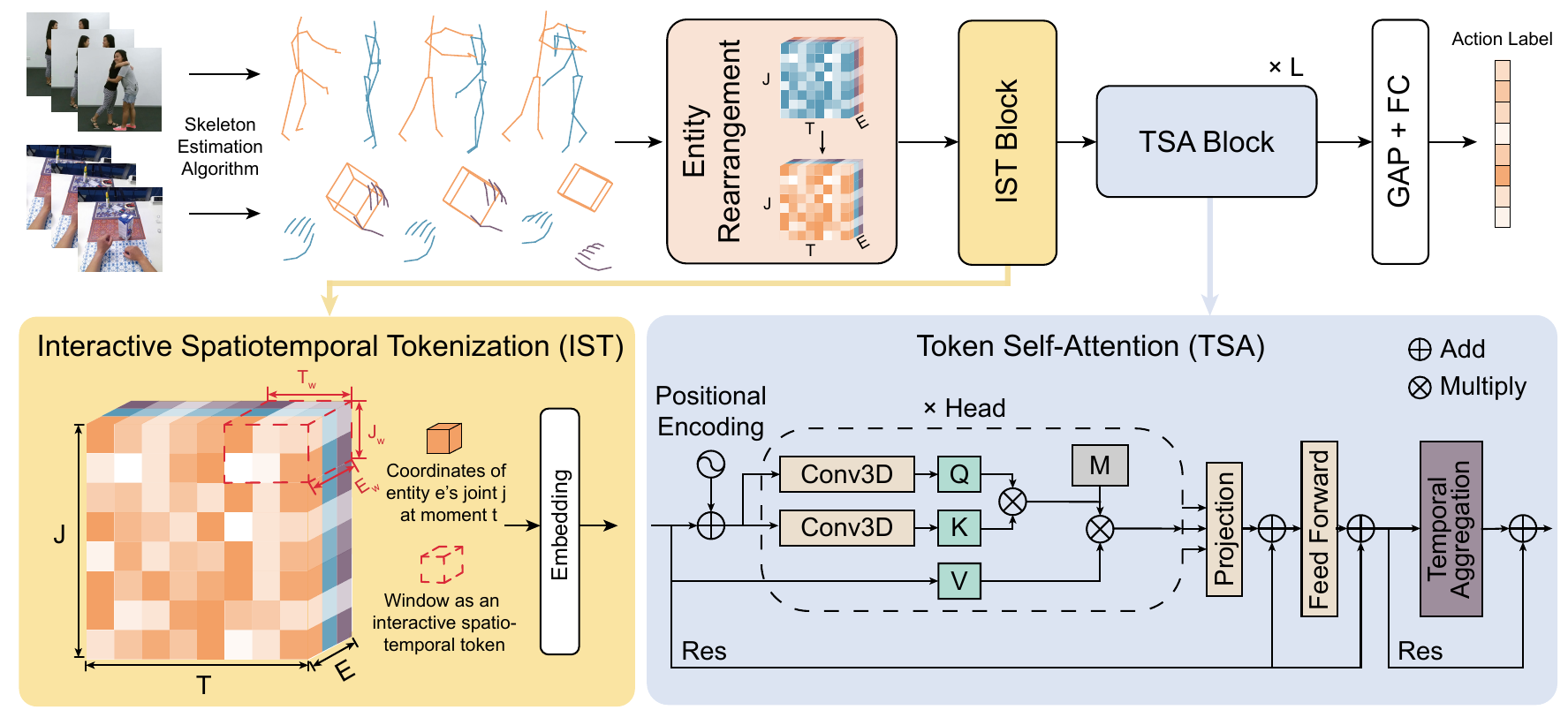}   
    \end{center}
    \vspace{-1.2em}
    \caption{The overall architecture of the proposed ISTA-Net for skeleton-based general interactive action recognition.}
    \label{method}
    \vspace{-1.2em}
\end{figure*}

\section{RELATED WORK}
\subsection{Action Recognition}
Most skeleton-based action recognition methods focus on developing effective architectures to recognize individual actions. Early approaches\cite{Co-LSTM2016,ST-LSTM2016, GCA2017, VA-LSTM2017, 2s-GCA2018} adopted RNN or LSTM to model long-term context of skeleton sequences. Then many models based on Graph Convolution Network (GCN) were proposed\cite{ST-GCN2018,AS-GCN2019, 2s-AGCN2019,MS-G3D2020,CTR-GCN2021,LST2022,tcagcn2022,hdgcn2022,InfoGCN2022}. To facilitate modeling channel-wise topologies, CTR-GCN \cite{CTR-GCN2021} learns a shared topology for all channels and refines it for each channel. InfoGCN\cite{InfoGCN2022} adopts a novel learning objective to learn compact latent representations. Recent works explored the protential to introduce self-attention mechanism into skeleton spatiotemporal modeling\cite{dstanet2020,STSA-Net2023}. For instance, STSA-Net\cite{STSA-Net2023} adopted a spatiotemporal segments encoding strategy to fuse joint relations between frames. 

\subsection{Interactive Action Recognition}
Recently-proposed interactive action recognition models \cite{H+O2019,H2O_TA-GCN2021,LSTM-IRN2022,igformer2022} capture interactions based on specially-designed modules according to subject priors. TA-GCN \cite{H2O_TA-GCN2021} models the hand-to-object relationship with a topology-aware graph convolutional network, in which prior graph dependencies of hands are predefined. For person-to-person mutual actions, LSTM-IRN \cite{LSTM-IRN2022} adopts relational reasoning over the different relationships between the human joints during interactions. IGFormer \cite{igformer2022} is the first to adopt Transformer-based architecture and leverages prior knowledge on human body structure to design co-attention mechanism for interactions. Different from above methods, our method utilizes Interactive Spatiotemporal Tokens as early fusions for modeling interactive spatiotemporal features, which also allows ISTA-Net to be able to handle various interactions, such as person-to-person, hand-to-hand, and hand-to-object interactions, with no need to manually predefine adjacency based on subject-type-specific prior knowledge.

\section{ISTA-Net}

Architecture of our proposed Interactive Spatiotemporal Token Attention Network is presented in Fig. \ref{method}. The input is an interactive action, which can be constituted by different types of entity. Firstly, ISTA-Net performs Entity Rearrangement in training to maintain the equivalence of unordered subjects. Subsequently the skeleton tensor gets tokenized by a 3D sliding window. Then the interactive Spatiotemporal Tokens are fed to \textit{L} Token Self-Attention Blocks to learn token-level interdependency. Prediction is finally made through Global Average Pooling (GAP) along ISTs following with a fully connected (FC) layer.

\subsection{Interactive Spatiotemporal Tokenization for Interactive Skeleton Sequences}

An important aspect of ISTA-Net is the design of attention tokens that represent interactive spatiotemporal local features for interactive skeleton sequences. We propose a general solution to represent motion of multiple skeletons including diverse subjects, without the assumption that priors of each interacting entity are already known.

Suppose that there are $E$ interactive entities performing an interaction over a period of time $T$, and each entity contains $J$ joints. Depending on whether 2D or 3D skeletons are estimated, the coordinate dimension $C$ can be 2 or 3. Thereby the input skeleton sequence is defined as $X_{input} \in \mathbb{R}^{C\times T\times J \times E}$. In comparison to individual actions, interactive actions have an additional dimension $E$ representing interactive entity parts or joints, which must be taken into consideration when tokenizing the skeletal data.

Our solution is to use non-overlapping 3D windows to obtain Interactive Spatiotemporal Tokens. This step is called Interactive Spatiotemporal Tokenization (IST) Block. Given a window $W$ of size $T_{w}\times J_{w} \times E_{w}$, it slides along temporal, spatial and interactive dimensions, partitioning the input data in a non-overlapping manner. Therefore, the input of size $C\times T\times J \times E$ is divided into $U = \lceil T/T_{w}\rceil \times \lceil J/J_{w}\rceil \times \lceil E/E_{w}\rceil$ patches of size $C\times T_{w}\times J_{w} \times E_{w}$ in total, which is illustrated as follows:
\begin{equation}
    X_{w} = IST(X_{input}, W),
\end{equation}
where $W \in \mathbb{R}^{T_{w}\times J_{w} \times E_{w}}$ and $X_{w} \in \mathbb{R}^{C\times T_{w}\times (J_{w}\times E_{w})\times U}$.

The tokens $X_{w}$ can be viewed in $\mathbb{R}^{(C\times  T_{w}\times J_{w}\times E_{w})\times U}$, which could be illustrated more clearly as the standard Transformer input format. However, in this case, we retain the coordinate dimension $C$ and temporal dimension $T_{w}$ for downsampling and temporal aggregation in later stages.

In some cases, such as in the $T$ channel, the input size $T$ may not be evenly divisible by the window size $T_w$. In such cases, parts of the original tensor should be replicated and padded along the $T$ dimension to create a new tensor of size $T'$ in time channel, where $T_w$ is an aliquot part of $T'$.

To enrich the representation in coordinates, a 3D $1\times 1 \times 1$ convolution is employed to extend the coordinate dimension from $C$ to $C'$, which could be formulated as
\begin{equation}
    X_{w}' = Conv3D_{(1\times 1 \times 1)}(X_{w}),
\end{equation}
where $X_{w}' \in \mathbb{R}^{C'\times T_{w}\times (J_{w}\times E_{w})\times U}$.

The 3D convolution operation, followed by the batch normalization and an activation function, serves as the embedding layer for interactive spatiotemporal tokens. Finally these tokens $X_{ist}$ are fed to several Multi-head Self-attention Blocks to learn high-level cross frame, joint and subject representations.

\subsection{Entity Rearrangement}
When partitioning ISTs as well as encoding positional information, the presence of a strict entity ordering can impede learning's ability to generalize to more cases. Specifically, for interactive entities engaged in mutual actions, some are semantically ordered and not interchangeable (e.g. left hand, right hand and object), while others are unordered and interchangeable (e.g. persons). The semantic equivalence of mutual subjects implies that the unordered entities are permutation-invariant. They can be arranged in any order while still representing the same interactive action.

This observation inspires us a simple yet effective way to eliminate the orderliness of interchangeable entities. Given the input skeleton sequence of size $C\times T\times J \times E$, we first divide it into $E$ parts along interactive dimension, obviously each of which represents the joint motion of one subject:
\begin{equation}
    [X_{1}, X_{2},\cdots, X_{i},\cdots, X_{E}] = Split(X_{input}),
\end{equation}
where $[1, 2,\cdots, i,\cdots, E]$ are indexes of the positional order along interactive dimension.

We could rearrange the original $X_{input}$ as follows:
\begin{equation}
    \tilde{X}_{input} = Concat([X_{v_1}, X_{v_2},\cdots, X_{v_i},\cdots, X_{v_E}]),
\end{equation}
where $[v_{1}, v_{2},\cdots, v_{i},\cdots, v_{E}]$ is an arbitrary arrangement of indexes $[1, 2,\cdots, i,\cdots, E]$.

\begin{algorithm}[t]
	\renewcommand{\algorithmicrequire}{\textbf{Input:}}
	\renewcommand{\algorithmicensure}{\textbf{Output:}}
	\caption{Interactive Spatiotemporal Tokenization with Entity Rearrangement}  
	\label{tokenize}
	\begin{algorithmic}[1]
 	    \Require The input skeleton sequence $X_{input} \in \mathbb{R}^{C\times T\times J \times E}$, the 3D sliding windows $W \in \mathbb{R}^{T_{w}\times J_{w} \times E_{w}}$, the ER boolean variable $\eta \in [0, 1]$, the embedding dimension $C'$, and the negative slope $\gamma$ for LeakyReLu.
		\If{$\eta$}
            \State{$[v_1, v_2, \cdots, v_E]\leftarrow$ Randperm$([1,2,\cdots,E])$}
		\State{$\tilde{X}_{input} \leftarrow X_{input}[:,:,:,[v_1, v_2, \cdots, v_E]]$}
		\Else
		\State{$\tilde{X}_{input}\leftarrow X_{input}$}
		\EndIf
        \State{$pad_{T1}, pad_{J1}, pad_{E1}\leftarrow 0$}
        \State{$pad_{T2}\leftarrow$ mod$((T_w - $mod$(T, T_w)), T_w)$}
        \State{$pad_{J2}\leftarrow$ mod$((J_w - $mod$(J, J_w)), J_w)$}
        \State{$pad_{E2}\leftarrow$ mod$((E_w - $mod$(E, E_w)), E_w)$}
        \State{$\tilde{X}_{input}\leftarrow$ pad$(\tilde{X}_{input},(pad_{E1}, pad_{E2}, pad_{J1}, pad_{J2},$}
        \Statex{$pad_{T1}, pad_{T2}))$}
        \State{$U\leftarrow \lceil T / T_w\rceil \times \lceil J / J_w\rceil \times \lceil E / E_w\rceil$}
        \State{$\tilde{X}_{input}\leftarrow \tilde{X}_{input}$.view$(C, T_w, \lceil T / T_w\rceil, J, \lceil J / J_w\rceil,$}
        \Statex{$E, \lceil E / E_w\rceil)$}
        \State{$\tilde{X}_{input}\leftarrow \tilde{X}_{input}$.permute$(0, 1, 3, 5, 2, 4, 6)$}
        \State{$X_{w}\leftarrow \tilde{X}_{input}$.view$(C, T_w, J_w\times E_w, U)$}
        \State{$X_{w}' \leftarrow$ Conv3D$_{(1\times 1 \times 1)}(X_{w}, (C, C'))$}
        \State{$X_{ist}\leftarrow$ LeakyReLu$($BatchNorm3D$(X_{w}'), \gamma)$}
        
		\State{\Return{$X_{ist}\in \mathbb{R}^{C'\times T_{w}\times (J_{w}\times E_{w})\times U}$}}
	\end{algorithmic}
\end{algorithm}

The complete process of our proposed Interactive Spatiotemporal Tokenization with Entity Rearrangement is illustrated in \textbf{Algorithm \ref{tokenize}}. Line 1-6 refer to ER. Line 7-11 refer to tensor padding. Line 12-15 refer to tokenziation using 3D windows. Line 16-17 represent embedding layers.

During each training epoch, an input permutation $\tilde{X}_{input}$ is selected, while in validation and testing, the original input $X_{input}$ is used. The total number of possible permutations for entities is $E!$, indicating that each permutation has a probability of $1/E!$ to be chosen as input. A theoretical concern is that the factorial increase in the number of samples may lead to non-convergent training. However, in practice, $E$ is typically small, since in most cases, there are not many mutual subjects in a single interactive action.

\begin{table*}[t]
	\centering
	\caption{Statistics of Interactive Action Recognition Datasets}
	\vspace{-0.7em}
	\label{dataset}
        \begin{threeparttable}
	\begin{tabular}{l|c|c|c|c|c|c|c|c|c}
		\hline
            \multirow{2}{*}{Datasets}&\multicolumn{3}{c|}{Annotation}&\multirow{2}{*}{\#Actions}&\multirow{2}{*}{\#Joints}&\multirow{2}{*}{\#Segments}&\multirow{2}{*}{Avg. Valid Frames}&\multirow{2}{*}{\#Entities}&\multirow{2}{*}{\#Participants}\\	
            \cline{2-4}
            &Persons&Hands&Object&&&&&&\\
		\hline
            NTU Mutual\cite{NTU120}&\Checkmark&&&26&25&24,732&59.36&2&106\\	
            SBU\cite{SBU}&\Checkmark&&&8&15&282&36.53&2&7\\	
            H2O\cite{H2O_TA-GCN2021}&&\Checkmark&\Checkmark&36&21&933&97.29&3&4\\	
            Assembly101\cite{Assembly101}&&\Checkmark&&1,380&21&85,252&105.91&2&53\\	
            \hline
	\end{tabular}
       \end{threeparttable}
\end{table*}

\begin{figure*}[t]
    \begin{center}
    \includegraphics[width=16cm]{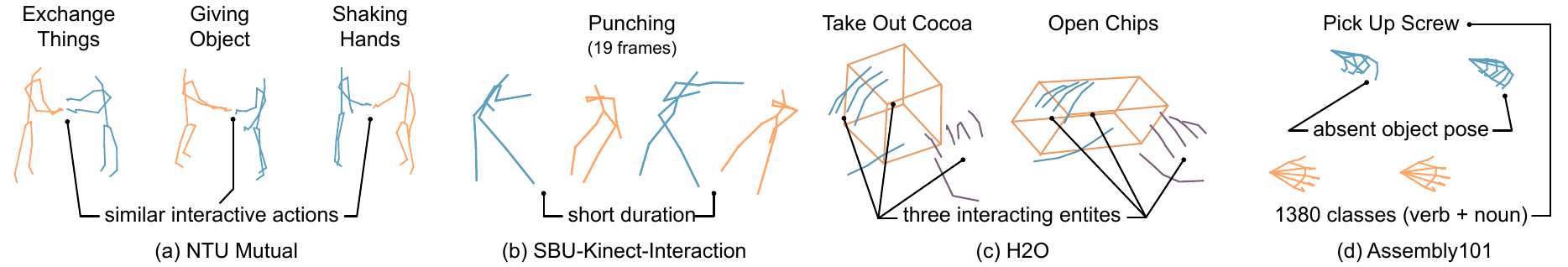}   
    \end{center}
    \vspace{-1.5em}
    \caption{Difficulties of interactive action recognition of diverse entities in four datasets.}
    \label{datasetviz}
    \vspace{-0.8em}
\end{figure*}

\subsection{Token Self-Attention Blocks}
To model the spatial, temporal, and interactive relationships simultaneously, our architecture incorporates a multi-head self-attention mechanism instead of a graph-convolution-based design. Unlike many GCNs, which require manual definition of an adjacency list for every joint based on prior knowledge of the physical connections between joints, our proposed architecture omits this tedious step for diverse action subjects. This also provides a unified approach to recognize interactive actions of diverse subjects .

Our proposed ISTA-Net consists of $L$ Token Self-Attention Blocks. Similar to standard multi-head self-attention, the input $X_{L_{i-1}}$ transforms to multiple sets of query $Q$, key $K$ and value $V$ as follows:  
\begin{equation}
    Q = Conv3D_{(1\times 1 \times 1)}(X_{L_{i-1}}+PE(X_{L_{i-1}})),
\end{equation}
\begin{equation}
    K = Conv3D_{(1\times 1 \times 1)}(X_{L_{i-1}}+PE(X_{L_{i-1}})),
\end{equation}
\begin{equation}
    V = X_{L_{i-1}},
\end{equation}
where positional encoding implemented with circular functions is denoted as $PE(\cdot)$. The number of sets, namely heads, is denoted as $H$.

Self-attention scores $X^{h}_{L_i}$ of the $h$-th head could be calculated as the following formula:
\begin{equation}
    X^{h}_{L_i} = (\alpha \tanh{(\frac{QK^{T}}{\sqrt{C_{\beta}}})} + M)V,
\end{equation}
where $QK^{T}$ is divided by the square root of the feature length $C_{\beta}=T_{w}\times J_{w} \times E_{w} \times C_{L_{i}-qkv}$. A trainable regularized matrix $M \in \mathbb{R}^{U\times U}$ is added to the normalized attention map with a trainable balanced factor $\alpha$, which can benefit correlation learning\cite{dstanet2020,STSA-Net2023}. All scores $X^{h}_{L_i}$ of $H$ heads are concatenated to get $X^{H}_{L_i}$. 

In some TSA Blocks, the $C_{L_{i-1}}$ dimension is doubled to downsample the features ($C_{L_{i}}=2\times C_{L_{i-1}}$), while in the others it remains the same ($C_{L_{i}}=C_{L_{i-1}}$):
\begin{equation}
    \hat{X}_{L_i} = Conv3D_{(1\times 1 \times k_u)}(X^{H}_{L_i}),
\end{equation}
\begin{equation}
    \acute{X}_{L_i} = Conv3D_{(1\times 1 \times 1)}(\hat{X}_{L_i}+X^{Res}_{L_{i}}) + X^{Res}_{L_{i}},
\end{equation}
where a 3D $1\times 1 \times 1$ convolution with residual connections implements the feed forward network (FFN).

The last component is the Temporal Aggregation (TA) layer. Previous researches\cite{tcagcn2022,hdgcn2022} indicate that feature aggregation along temporal channel is effective for modeling actions. In contrast to those methods, the proposed ISTA-Net uses 3D convolution with kernel sizes larger than 1 in the temporal dimension ($k_t>1$) to aggregate sequence features:
\begin{equation}
    {X}_{L_i} = Conv3D_{(k_t\times 1 \times 1)}(\acute{X}_{L_i}) + \acute{X}^{Res}_{L_i},
\end{equation}
which is followed by a residual connection $\acute{X}^{Res}_{L_i}$.

\begin{table*}[t]
        \renewcommand\arraystretch{1.3}
	\centering
	\caption{Comparisons of Action Recognition Methods on Four Different Interactive Action Datasets}
	\vspace{-0.7em}
	\label{sota}
        \resizebox{\textwidth}{!}{
        \begin{threeparttable}
        \resizebox{\textwidth}{!}{
	\begin{NiceTabular}{c|l|c|c|C{22mm}|C{22mm}|c|c}[colortbl-like]
		\hline
            \multirow{2}{*}{Type}&
		\multicolumn{1}{c}{\multirow{2}{*}{Methods\tnote{1}}}&
            \multirow{2}{*}{Year}&
            \multirow{2}{*}{SBU(\%)}&
            \multicolumn{2}{c|}{NTU RGB+D 120 - 26 Mutual Actions(\%)}&
            \multirow{2}{*}{H2O(\%)}&
            \multirow{2}{*}{Assembly101(\%)}\\
            \cline{5-6}

            &
		&
            &
            &
            X-Sub&
            X-Set&
            &
            \\

            \hline

            \rowcolor{myYellow!15}
            &
            Co-LSTM\cite{Co-LSTM2016}&
            AAAI 2016&
            90.40&
            -&
            -&
            -&
            -\\

            \rowcolor{myYellow!15}
            &
            ST-LSTM\cite{ST-LSTM2016}&
            ECCV 2016&
            93.30&
            63.00&
            66.60&
            -&
            -\\

            \rowcolor{myYellow!15}
            &
            GCA\cite{GCA2017}&
            CVPR 2017&
            -&
            70.60&
            73.70&
            -&
            -\\

            \rowcolor{myYellow!15}
            &
            VA-LSTM\cite{VA-LSTM2017}&
            ICCV 2017&
            97.20&
            -&
            -&
            -&
            -\\

            \rowcolor{myYellow!15}
            &
            2s-GCA\cite{2s-GCA2018}&
            TIP 2018&
            94.90&
            73.00&
            73.30&
            -&
            -\\

            \rowcolor{myYellow!15}
            &
            H+O\cite{H+O2019}&
            CVPR 2019&
            -&
            -&
            -&
            68.88&
            -\\

            \rowcolor{myYellow!15}
            \multirow{-7}{*}{LSTM}&
            LSTM-IRN\cite{LSTM-IRN2022}&
            TMM 2022&
            98.20&
            77.70&
            79.60&
            -&
            -\\

            \hline

            \rowcolor{myRed!20}
            &
            ST-GCN\cite{ST-GCN2018}&
            AAAI 2018&
            -&
            78.90&
            76.10&
            73.86&
            -\\

            \rowcolor{myRed!20}
            &
            AS-GCN\cite{AS-GCN2019}&
            CVPR 2019&
            -&
            82.90&
            83.70&
            -&
            -\\

            \rowcolor{myRed!20}
            &
            2s-AGCN\cite{2s-AGCN2019}&
            CVPR 2019&
            -&
            -&
            -&
            -&
            26.70\\

            \rowcolor{myRed!20}
            &
            MS-G3D\cite{MS-G3D2020}&
            CVPR 2020&
            -&
            -&
            -&
            -&
            26.86 ($\pm$0.11)\\

            \rowcolor{myRed!20}
            &
            CTR-GCN\cite{CTR-GCN2021}&
            ICCV 2021&
            -&
            89.32 ($\pm$0.06)&
            90.19 ($\pm$0.17)&
            -&
            26.25 ($\pm$0.81) \\

            \rowcolor{myRed!20}
            &
            TA-GCN\cite{H2O_TA-GCN2021}&
            ICCV 2021&
            -&
            -&
            -&
            79.25&
            -\\

            \rowcolor{myRed!20}
            &
            LST\cite{LST2022}&
            arXiv 2022&
            -&
            89.27 ($\pm$0.23)&
            90.60 ($\pm$0.13)&
            -&
            -\\

            \rowcolor{myRed!20}
            &
            TCA-GCN\cite{tcagcn2022}&
            arXiv 2022&
            -&
            88.37 ($\pm$0.38)&
            89.30 ($\pm$0.34)&
            -&
            -\\

            \rowcolor{myRed!20}
            &
            HD-GCN\cite{hdgcn2022}&
            arXiv 2022&
            -&
            88.25 ($\pm$0.44)&
            90.08 ($\pm$0.12)&
            -&
            -\\

            \rowcolor{myRed!20}
            \multirow{-10}{*}{GCN}&
            InfoGCN\cite{InfoGCN2022}&
            CVPR 2022&
            -&
            90.22 ($\pm$0.13)&
            91.13 ($\pm$0.16)&
            -&
            25.63 ($\pm$0.21)\\

            \hline

            \rowcolor{myBlue!10}
            &
            DSTA-Net\cite{dstanet2020}&
            ACCV 2020&
            -&
            88.92 ($\pm$0.26)&
            90.10 ($\pm$0.24)&
            -&
            -\\

            \rowcolor{myBlue!10}
            &
            STSA-Net\cite{STSA-Net2023}&
            Neurocomputing 2023&
            -&
            90.20 ($\pm$0.16)&
            90.97 ($\pm$0.25)&
            -&
            -\\

            \rowcolor{myBlue!10}
            &
            IGFormer\cite{igformer2022}&
            ECCV 2022&
            98.40&
            85.40&
            86.50&
            -&
            22.33 ($\pm$0.14)\\

            \rowcolor{myBlue!10}
            \multirow{-4}{*}{Transformer}&
            ISTA-Net (Ours)&
            2023&
            \textbf{98.51 ($\pm$1.47)}&
            \textbf{90.56 ($\pm$0.08)}&
            \textbf{91.72 ($\pm$0.30)}&
            \textbf{89.09 ($\pm$1.21)}&
            \textbf{28.01 ($\pm$0.06)}\\
		
		\hline
	\end{NiceTabular}
        }
        \begin{tablenotes}
         \item[1] This table reports the averaged top-1 accuracy in several seed initializations, along with the standard deviation in brackets. Statisics without brackets are cited from \cite{igformer2022, H2O_TA-GCN2021, Assembly101}.
       \end{tablenotes}
       \end{threeparttable}
        }
        \vspace{-1.0em}
\end{table*}

\section{EXPERIMENTS}
\subsection{Datasets}

\textbf{NTU RGB+D 120}\cite{NTU120}, the extension version of \textbf{NTU RGB+D}\cite{NTU60}, is a widely-used action recognition dataset. It provides 114,480 samples of 120 human actions. In our experiments we focus on a subset of NTU RGB+D 120 Dataset, which consists of 26 kinds of mutual actions (named \textbf{NTU Mutual}, for short). 

\textbf{SBU-Kinect-Interaction}\cite{SBU} is a human activity
dataset that depicts person-to-person interactions. It includes eight interactions, with RGB+D videos and extracted skeletons.

\textbf{H2O}\cite{H2O_TA-GCN2021} is the first dataset constructured for egocentric 3D interaction recognition. With 3D pose of both hands and pose of manipulated objects, H2O dataset facilitates hand-to-hand and hand-to-object interactions understanding.

\textbf{Assembly101}\cite{Assembly101} is a large procedural activity dataset. 3D hand poses are provided to advance 3D interaction recognition from egocentric views. It's a tough task due to the dataset's complexity, which includes over 1,300 fine-grained classes of hand-to-object interactions. Each class consists of a single verb and an object that is manipulated. Additionally, the absence of object poses adds another layer of difficulty to judging the interactive actions.

Statistics and difficulties of these datasets are summarized in Table \ref{dataset} and Fig. \ref{datasetviz}. For evaluation on NTU Mutual, we employ the Cross-subject (X-Sub) and Cross-set (X-Set) criteria \cite{NTU120}, using only the joint modality to ensure fair comparisons without fusion. For SBU, the suggested 5-fold cross validation approach \cite{SBU} is adopted. For H2O and Assembly101, we follow the training, validation, and test splits described in \cite{H2O_TA-GCN2021} and \cite{Assembly101}, respectively.

\subsection{Implementation Details}
All of our experiments are conducted on a machine equipped with four GeForce RTX 3070 GPUs and CUDA version 11.4. For training on NTU Mutual dataset, SGD optimizer is used with Nesterov momentum of 0.9, a initial learning rate of 0.1 and a decay rate 0.1. Window size is set to [20, 1, 2]. Cross entropy is used as loss function with label smoothing factor 0.1 and temperature factor 1.0. Batch size is 32. Each training process was terminated after 110 epochs. Parameters for the other datasets might be different. Please refer to the configurations in our Github repository. 

\subsection{Comparison with Related Methods}

\begin{figure*}[t]
    \begin{center}
    \includegraphics[width=18cm]{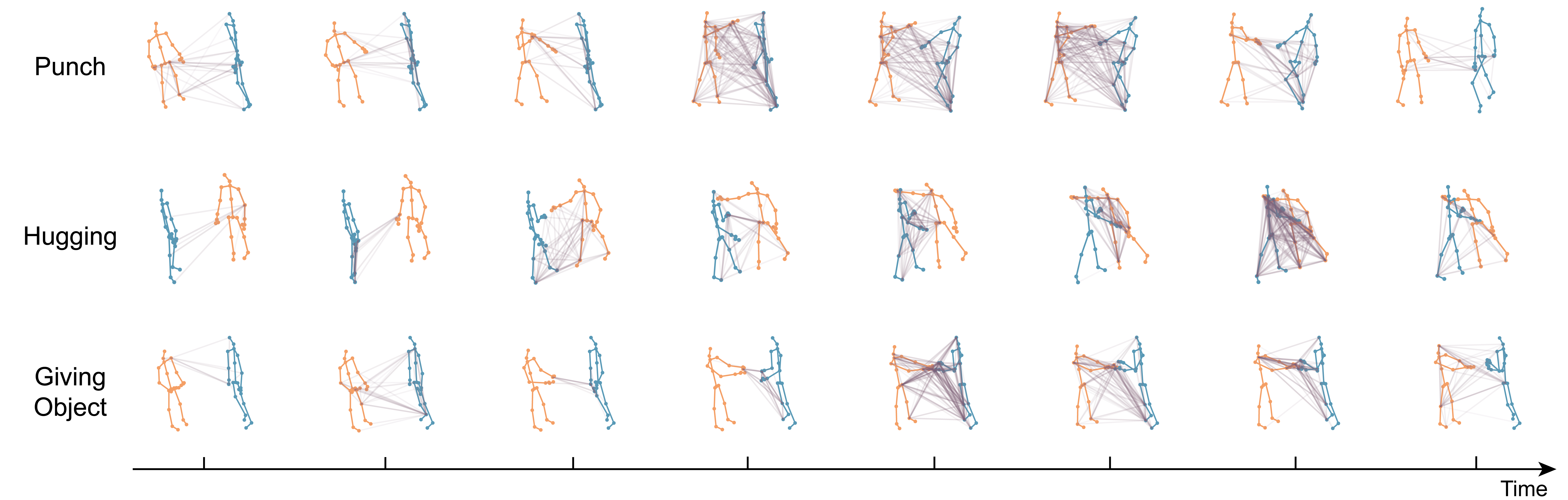}
    \end{center}
    \vspace{-1.2em}
    \caption{Visualization of the learnt interactive relations restored from the last TSA Block. The attentive weights are visualized to illustrate the important body parts involved in recognizing different interactive actions. Specifically, ISTA-Net recognizes the \textit{Punch} action through attentions on the attacker's hands and the victim's limbs. The \textit{Hugging} action is recognized through attentions on the approaching and contacting body parts. The \textit{Giving Object} action is recognized through attentions on the hands.}
    \label{Viz}
    \vspace{-0.3em}
\end{figure*}

Table \ref{sota} reports the experimental results on NTU Mutual, SBU, H2O and Assembly101 datasets. The proposed ISTA-Net achieves state-of-the-art performance compared with other traditional action recognition and interactive action recognition methods. Benefitting from the proposed ISTs, TSA Blocks and ER, ISTA-Net outperforms many LSTM-, GCN-, and Transformer-based action recognition methods. ISTA-Net achieves 5.16\%, 5.22\%, 0.11\% and 5.68\% gains over the most related interactive action recognition method, IGFormer\cite{igformer2022}, on NTU Mutual X-Sub, X-Set, SBU and Assembly101. ISTA-Net also outperforms InfoGCN\cite{InfoGCN2022} by 0.34\% and 0.59\% on NTU Mutual, TA-GCN\cite{H2O_TA-GCN2021} by 9.84\% on H2O, and MS-G3D\cite{MS-G3D2020} by 1.15\% on Assembly101. Observed from the results, our ISTA-Net also show its superiority and adaptability to diverse interactive entities. Fig. \ref{Viz} visualizes the learnt attention in the last TSA Block, which verifies the effectiveness of ISTA-Net when modeling interactive actions.

\subsection{Ablation Study}

\begin{table}[t]
	\centering
	\caption{Comparison of Ways to Fuse Interactive Relations}
	\vspace{-0.7em}
	\label{ablation1}
	\begin{tabular}{l|c|c}
            \hline
                & NTU Mutual X-Sub (\%) & $\Delta$ (\%) \\
		\hline
                \textbf{IST} & \textbf{90.56 ($\pm$0.08)} & - \\
                Co-attention & 89.78 ($\pm$0.16) & -0.78\\
                Late Fusion & 88.79 ($\pm$0.24) & -1.77\\
                Coordinate Concat & 88.14 ($\pm$0.01) & -2.42\\
            \hline
	\end{tabular}
        \vspace{-0.8em}
\end{table}

\begin{table}[t]
	\centering
	\caption{Effectiveness of Entity Rearrangements}
	\vspace{-0.7em}
	\label{ablation2}
        \resizebox{\columnwidth}{!}{
	\begin{tabular}{l|c|c|c|c}
            \hline
                & NTU Mutual X-Sub (\%) & $\Delta$ (\%) & SBU (\%) & $\Delta$ (\%)  \\
		\hline
                \textbf{w/ ER} & \textbf{90.56 ($\pm$0.08)} & - & \textbf{98.51 ($\pm$1.47)} & - \\
                w/o ER & 90.25 ($\pm$0.18) &-0.31& 94.90 ($\pm$3.73) & -3.61 \\
            \hline
	\end{tabular}
        }
        \vspace{-1.0em}
\end{table}

\begin{table}[t]
	\centering
	\caption{Effectiveness of Temporal Aggregation}
	\vspace{-0.7em}
	\label{ablation3}
	\begin{tabular}{l|c|c}
            \hline
                & NTU Mutual X-Sub (\%) & $\Delta$ (\%)\\
		\hline
                \textbf{w/ TA} & \textbf{90.56 ($\pm$0.08)} & - \\
                w/o TA & 87.45 ($\pm$0.14)& -3.11\\
            \hline
	\end{tabular}
        \vspace{-0.8em}
\end{table}

\begin{table}[t]
	\centering
	\caption{Performances Using Different Input Frame Lengths}
	\vspace{-0.7em}
	\label{ablation4}
	\begin{tabular}{l|c|c}
            \hline
               \#Frames & NTU Mutual X-Sub (\%) & $\Delta$ (\%)\\
		\hline
                180 & 90.09 ($\pm$0.05) & -0.47\\
                120 & \textbf{90.56 ($\pm$0.08)} & - \\
                60 & 90.42 ($\pm$0.20) & -0.14\\
                30 & 89.42 ($\pm$0.10) & -1.14\\
            \hline
	\end{tabular}
        \vspace{-1.0em}
\end{table}

\begin{table}[t]
	\centering
	\caption{Performances Using Different Window Sizes}
	\vspace{-0.7em}
	\label{ablation5}
	\begin{tabular}{l|c|c}
            \hline
               Window Size & NTU Mutual X-Sub (\%) & $\Delta$ (\%)\\
		\hline
                $[20, 1, 2]$ & \textbf{90.56 ($\pm$0.08)} & - \\
            \hline
                $[40, 1, 2]$ & 89.98 ($\pm$0.23)& -0.58\\
                $[10, 1, 2]$ & 90.13 ($\pm$0.18)& -0.43\\
            \hline
                $[20, 2, 2]$ & 90.31 ($\pm$0.16)& -0.25\\
                $[20, 5, 2]$ & 89.61 ($\pm$0.09)& -0.95\\
            \hline
                $[20, 1, 1]$ & 90.19 ($\pm$0.08)& -0.37\\
            \hline
	\end{tabular}
        \vspace{-1.0em}
\end{table}

\textbf{Comparison of Ways to Fuse Interactive Relations.} We compare four approaches to model the interactive relations of spatiotemporal features. The first approach, called \textit{Late Fusion}, is widely used in traditional action recognition methods when adapting to interactive skeletons. In \textit{Late Fusion}, interactions are only modeled in the classification head. The second one, \textit{Co-attention}, employs weight-shared dual-branch self-attention blocks. In each block, $K$ and $V$ are got from the previous block in this branch, while $Q$ is obtain from the other branch. The third approach, \textit{Coordinate Concat}, directly concatenates entity features along coordinate dimension. The last one is our proposed \textit{IST}, which fuses interactive features during early tokenization. Compared to the others, an additional dimension $E$ is extended in this method. Table \ref{ablation1} demonstrates that \textit{IST} outperforms the other approaches by 1.77\%, 0.78\% and 2.42\%.

\textbf{Effectiveness of Entity Rearrangement.} We explore the effectiveness of Entity Rearrangement by removing this step. As reported in Table \ref{ablation2}, the performance declined on the relatively larger NTU Mutual dataset, and more significantly on the relatively smaller SBU dataset. This indicates that ER is beneficial for enhancing model generalization, particularly when training with small-scale data.

\textbf{Effectiveness of Temporal Aggregation.} To confirm the contributions made by Temporal Aggregation, we removed this step for comparison purposes. The results in Table \ref{ablation3} indicate that TA can effectively aggregate local temporal motion features in ISTs and improve recognition performance.

\textbf{Comparisons of Different Input Frame Lengths and Window Sizes.} We evaluate the influence of various input frame lengths and window sizes on the performance of ISTA-Net. On NTU dataset, 60 and 120 are the two most widely-adopted input frame lengths. To ensure fair comparisons, when taking different numbers of frames, window size is scaled accordingly in temporal dimension, thus keeping the number of ISTs unchanged. The results presented in Table \ref{ablation4} suggest that using 120 frames as input achieves the best performance, and adding more frames introduces additional noise. Table \ref{ablation5} shows that, given a fixed number of frames, a window size of $[20,1,2]$ leads to the optimal result, indicating joints can be modeled better at a fine-grained level.

\section{CONCLUSIONS}

This paper proposes Interactive Spatiotemporal Token Attention Network for general interactive action recognition, which does not require subject-type-specific graph prior knowledge to model diverse interacting entities. Our ISTA-Net consists of Interactive Spatiotemporal Tokenization Block and Token Self-Attention Blocks. By extending an additional entity dimension in attention tokens, our design can simultaneously and also effectively capture interactive and spatiotemporal correlations of interactive actions. Moreover, we introduce Entity Rearrangement to preserve the disorderliness of unordered subjects in Interactive Spatiotemporal Tokens. Our approach shows superior performance and adaptability on four benchmarks of interactive action recognition.

\section{ACKNOWLEDGEMENT}
This work was supported by the National Natural Science Foundation of China (Grant No. 62203476, No. 52105079).

\addtolength{\textheight}{-1.5cm}   



\normalem
\bibliographystyle{./IEEEtran}
\bibliography{Reference}

\begin{thebibliography}{10}
\providecommand{\url}[1]{#1}
\csname url@rmstyle\endcsname
\providecommand{\newblock}{\relax}
\providecommand{\bibinfo}[2]{#2}
\providecommand\BIBentrySTDinterwordspacing{\spaceskip=0pt\relax}
\providecommand\BIBentryALTinterwordstretchfactor{4}
\providecommand\BIBentryALTinterwordspacing{\spaceskip=\fontdimen2\font plus
\BIBentryALTinterwordstretchfactor\fontdimen3\font minus
  \fontdimen4\font\relax}
\providecommand\BIBforeignlanguage[2]{{%
\expandafter\ifx\csname l@#1\endcsname\relax
\typeout{** WARNING: IEEEtran.bst: No hyphenation pattern has been}%
\typeout{** loaded for the language `#1'. Using the pattern for}%
\typeout{** the default language instead.}%
\else
\language=\csname l@#1\endcsname
\fi
#2}}

\bibitem{9636389}
P.~Wang, J.~Liu, F.~Hou, D.~Chen, Z.~Xia, and S.~Guo, ``Organization and
  understanding of a tactile information dataset tacact for physical
  human-robot interaction,'' in \emph{2021 IEEE/RSJ International Conference on
  Intelligent Robots and Systems (IROS)}, 2021, pp. 7328--7333.

\bibitem{9636110}
N.~Feng, F.~Hu, H.~Wang, and Z.~Zhao, ``Hybrid graph convolutional networks for
  skeleton-based and eeg-based jumping action recognition,'' in \emph{2021
  IEEE/RSJ International Conference on Intelligent Robots and Systems (IROS)},
  2021, pp. 4156--4161.

\bibitem{zheng2017image}
Z.-H. Zheng, H.-T. Zhang, F.-L. Zhang, and T.-J. Mu, ``Image-based clothes
  changing system,'' \emph{Computational Visual Media}, vol.~3, pp. 337--347,
  2017.

\bibitem{DBLP:conf/iros/XingB22}
H.~Xing and D.~Burschka, ``Understanding spatio-temporal relations in
  human-object interaction using pyramid graph convolutional network,'' in
  \emph{2022 IEEE/RSJ International Conference on Intelligent Robots and
  Systems (IROS)}, 2022, pp. 5195--5201.

\bibitem{9636381}
A.~Roitberg, D.~Schneider, A.~Djamal, C.~Seibold, S.~Reiß, and
  R.~Stiefelhagen, ``Let’s play for action: Recognizing activities of daily
  living by learning from life simulation video games,'' in \emph{2021 IEEE/RSJ
  International Conference on Intelligent Robots and Systems (IROS)}, 2021, pp.
  8563--8569.

\bibitem{7554295}
J.~Zhao, Y.~Ma, J.~Dong, and S.~Hou, ``Interactive mechanical arm control
  system based on kinect,'' in \emph{2016 35th Chinese Control Conference
  (CCC)}, 2016, pp. 5976--5981.

\bibitem{9636107}
D.~Zhang, N.~A. Vien, M.~Van, and S.~McLoone, ``Non-local graph convolutional
  network for joint activity recognition and motion prediction,'' in \emph{2021
  IEEE/RSJ International Conference on Intelligent Robots and Systems (IROS)},
  2021, pp. 2970--2977.

\bibitem{9636553}
H.~Xing, Y.~Xue, M.~Zhou, and D.~Burschka, ``Robust event detection based on
  spatio-temporal latent action unit using skeletal information,'' in
  \emph{2021 IEEE/RSJ International Conference on Intelligent Robots and
  Systems (IROS)}, 2021, pp. 2941--2948.

\bibitem{NTU120}
J.~Liu, A.~Shahroudy, M.~Perez, G.~Wang, L.-Y. Duan, and A.~C. Kot, ``Ntu rgb+d
  120: A large-scale benchmark for 3d human activity understanding,''
  \emph{IEEE Transactions on Pattern Analysis and Machine Intelligence},
  vol.~42, no.~10, pp. 2684--2701, 2020.

\bibitem{cad2009}
W.~Choi, K.~Shahid, and S.~Savarese, ``What are they doing? : Collective
  activity classification using spatio-temporal relationship among people,'' in
  \emph{2009 IEEE 12th International Conference on Computer Vision Workshops,
  ICCV Workshops}, 2009, pp. 1282--1289.

\bibitem{8299578}
Y.~Zhang, C.~Cao, J.~Cheng, and H.~Lu, ``Egogesture: A new dataset and
  benchmark for egocentric hand gesture recognition,'' \emph{IEEE Transactions
  on Multimedia}, vol.~20, no.~5, pp. 1038--1050, 2018.

\bibitem{H2O_TA-GCN2021}
T.~Kwon, B.~Tekin, J.~St\"uhmer, F.~Bogo, and M.~Pollefeys, ``H2o: Two hands
  manipulating objects for first person interaction recognition,'' in
  \emph{Proceedings of the IEEE/CVF International Conference on Computer Vision
  (ICCV)}, October 2021, pp. 10\,138--10\,148.

\bibitem{H+O2019}
B.~Tekin, F.~Bogo, and M.~Pollefeys, ``H+o: Unified egocentric recognition of
  3d hand-object poses and interactions,'' in \emph{2019 IEEE/CVF Conference on
  Computer Vision and Pattern Recognition (CVPR)}, 2019, pp. 4506--4515.

\bibitem{LSTM-IRN2022}
M.~Perez, J.~Liu, and A.~C. Kot, ``Interaction relational network for mutual
  action recognition,'' \emph{IEEE Transactions on Multimedia}, vol.~24, pp.
  366--376, 2022.

\bibitem{igformer2022}
Y.~Pang, Q.~Ke, H.~Rahmani, J.~Bailey, and J.~Liu, ``Igformer: Interaction
  graph transformer for skeleton-based human interaction recognition,'' in
  \emph{Computer Vision -- ECCV 2022}, S.~Avidan, G.~Brostow, M.~Ciss{\'e},
  G.~M. Farinella, and T.~Hassner, Eds.\hskip 1em plus 0.5em minus 0.4em\relax
  Cham: Springer Nature Switzerland, 2022, pp. 605--622.

\bibitem{Co-LSTM2016}
W.~Zhu, C.~Lan, J.~Xing, W.~Zeng, Y.~Li, L.~Shen, and X.~Xie, ``Co-occurrence
  feature learning for skeleton based action recognition using regularized deep
  lstm networks,'' in \emph{Proceedings of the Thirtieth AAAI Conference on
  Artificial Intelligence}, ser. AAAI'16.\hskip 1em plus 0.5em minus
  0.4em\relax AAAI Press, 2016, p. 3697–3703.

\bibitem{ST-LSTM2016}
J.~Liu, A.~Shahroudy, D.~Xu, and G.~Wang, ``Spatio-temporal lstm with trust
  gates for 3d human action recognition,'' in \emph{Computer Vision -- ECCV
  2016}, B.~Leibe, J.~Matas, N.~Sebe, and M.~Welling, Eds.\hskip 1em plus 0.5em
  minus 0.4em\relax Cham: Springer International Publishing, 2016, pp.
  816--833.

\bibitem{GCA2017}
J.~Liu, G.~Wang, P.~Hu, L.-Y. Duan, and A.~C. Kot, ``Global context-aware
  attention lstm networks for 3d action recognition,'' in \emph{2017 IEEE
  Conference on Computer Vision and Pattern Recognition (CVPR)}, 2017, pp.
  3671--3680.

\bibitem{VA-LSTM2017}
P.~Zhang, C.~Lan, J.~Xing, W.~Zeng, J.~Xue, and N.~Zheng, ``View adaptive
  recurrent neural networks for high performance human action recognition from
  skeleton data,'' in \emph{2017 IEEE International Conference on Computer
  Vision (ICCV)}, 2017, pp. 2136--2145.

\bibitem{2s-GCA2018}
J.~Liu, G.~Wang, L.-Y. Duan, K.~Abdiyeva, and A.~C. Kot, ``Skeleton-based human
  action recognition with global context-aware attention lstm networks,''
  \emph{IEEE Transactions on Image Processing}, vol.~27, no.~4, pp. 1586--1599,
  2018.

\bibitem{ST-GCN2018}
S.~Yan, Y.~Xiong, and D.~Lin, ``Spatial temporal graph convolutional networks
  for skeleton-based action recognition,'' in \emph{Proceedings of the
  Thirty-Second AAAI Conference on Artificial Intelligence and Thirtieth
  Innovative Applications of Artificial Intelligence Conference and Eighth AAAI
  Symposium on Educational Advances in Artificial Intelligence}, ser.
  AAAI'18/IAAI'18/EAAI'18.\hskip 1em plus 0.5em minus 0.4em\relax AAAI Press,
  2018.

\bibitem{AS-GCN2019}
M.~Li, S.~Chen, X.~Chen, Y.~Zhang, Y.~Wang, and Q.~Tian, ``Actional-structural
  graph convolutional networks for skeleton-based action recognition,'' in
  \emph{2019 IEEE/CVF Conference on Computer Vision and Pattern Recognition
  (CVPR)}, 2019, pp. 3590--3598.

\bibitem{2s-AGCN2019}
L.~Shi, Y.~Zhang, J.~Cheng, and H.~Lu, ``Two-stream adaptive graph
  convolutional networks for skeleton-based action recognition,'' in \emph{2019
  IEEE/CVF Conference on Computer Vision and Pattern Recognition (CVPR)}, 2019,
  pp. 12\,018--12\,027.

\bibitem{MS-G3D2020}
Z.~Liu, H.~Zhang, Z.~Chen, Z.~Wang, and W.~Ouyang, ``Disentangling and unifying
  graph convolutions for skeleton-based action recognition,'' in \emph{2020
  IEEE/CVF Conference on Computer Vision and Pattern Recognition (CVPR)}, 2020,
  pp. 140--149.

\bibitem{CTR-GCN2021}
Y.~Chen, Z.~Zhang, C.~Yuan, B.~Li, Y.~Deng, and W.~Hu, ``Channel-wise topology
  refinement graph convolution for skeleton-based action recognition,'' in
  \emph{2021 IEEE International Conference on Computer Vision (ICCV)}, 2021,
  pp. 13\,359--13\,368.

\bibitem{LST2022}
W.~Xiang, C.~Li, Y.~Zhou, B.~Wang, and L.~Zhang, ``Language supervised training
  for skeleton-based action recognition,'' \emph{arXiv preprint
  arXiv:2208.05318}, 2022.

\bibitem{tcagcn2022}
\BIBentryALTinterwordspacing
S.~Wang, Y.~Zhang, M.~Zhao, H.~Qi, K.~Wang, F.~Wei, and Y.~Jiang,
  ``Skeleton-based action recognition via temporal-channel aggregation,'' 2022.
  [Online]. Available: \url{https://arxiv.org/abs/2205.15936}
\BIBentrySTDinterwordspacing

\bibitem{hdgcn2022}
\BIBentryALTinterwordspacing
J.~Lee, M.~Lee, D.~Lee, and S.~Lee, ``Hierarchically decomposed graph
  convolutional networks for skeleton-based action recognition,'' 2022.
  [Online]. Available: \url{https://arxiv.org/abs/2208.10741}
\BIBentrySTDinterwordspacing

\bibitem{InfoGCN2022}
H.-G. Chi, M.~H. Ha, S.~Chi, S.~W. Lee, Q.~Huang, and K.~Ramani, ``Infogcn:
  Representation learning for human skeleton-based action recognition,'' in
  \emph{2022 IEEE/CVF Conference on Computer Vision and Pattern Recognition
  (CVPR)}, 2022, pp. 20\,154--20\,164.

\bibitem{dstanet2020}
L.~Shi, Y.~Zhang, J.~Cheng, and H.~Lu, ``Decoupled spatial-temporal attention
  network for skeleton-based action-gesture recognition,'' in \emph{Computer
  Vision – ACCV 2020: 15th Asian Conference on Computer Vision, Kyoto, Japan,
  November 30 – December 4, 2020, Revised Selected Papers, Part V}.\hskip 1em
  plus 0.5em minus 0.4em\relax Berlin, Heidelberg: Springer-Verlag, 2020, p.
  38–53.

\bibitem{STSA-Net2023}
H.~Qiu, B.~Hou, B.~Ren, and X.~Zhang, ``Spatio-temporal segments attention for
  skeleton-based action recognition,'' \emph{Neurocomputing}, vol. 518, pp.
  30--38, 2023.

\bibitem{SBU}
K.~Yun, J.~Honorio, D.~Chattopadhyay, T.~L. Berg, and D.~Samaras, ``Two-person
  interaction detection using body-pose features and multiple instance
  learning,'' in \emph{2012 IEEE Computer Society Conference on Computer Vision
  and Pattern Recognition Workshops}, 2012, pp. 28--35.

\bibitem{Assembly101}
F.~Sener, D.~Chatterjee, D.~Shelepov, K.~He, D.~Singhania, R.~Wang, and A.~Yao,
  ``Assembly101: A large-scale multi-view video dataset for understanding
  procedural activities,'' in \emph{2022 IEEE/CVF Conference on Computer Vision
  and Pattern Recognition (CVPR)}, 2022, pp. 21\,064--21\,074.

\bibitem{NTU60}
A.~Shahroudy, J.~Liu, T.-T. Ng, and G.~Wang, ``Ntu rgb+d: A large scale dataset
  for 3d human activity analysis,'' in \emph{2016 IEEE Conference on Computer
  Vision and Pattern Recognition (CVPR)}, 2016, pp. 1010--1019.

\end{thebibliography}

\end{document}